\documentclass{sigchi}

\usepackage{balance} 
\usepackage{graphics}
\usepackage{times}   
\usepackage{url}     
\usepackage{color, colortbl}
\usepackage{textcomp}
\usepackage{placeins}

\makeatletter
\def\url@leostyle{%
  \@ifundefined{selectfont}{\def\UrlFont{\sf}}{\def\UrlFont{\small\bf\ttfamily}}}
\makeatother
\def\pprw{210mm}
\def\pprh{297mm}

\usepackage[pdftex]{hyperref}
\hypersetup{
pdftitle={Pupil: An Open Source Platform for Pervasive Eye Tracking and Mobile Gaze-based Interaction},
pdfauthor={Moritz Kassner, William Patera, Andreas Bulling},
pdfkeywords={Eye Movement, Mobile Eye Tracking, Wearable Computing, Gaze-based Interaction},
bookmarksnumbered,
pdfstartview={FitH},
colorlinks,
citecolor=green,
filecolor=black,
linkcolor=green,
urlcolor=blue,
breaklinks=true,
}

\newcommand{\figref}[1]{\hypersetup{linkcolor=red}\ref{#1}\hypersetup{linkcolor=black}}

\begin{document}

\title{Pupil: An Open Source Platform for Pervasive Eye Tracking and Mobile Gaze-based Interaction}

\numberofauthors{3}
\author{
  \alignauthor Moritz Kassner\\
    \affaddr\small{Pupil Labs UG}\\
    \affaddr\small{Berlin, Germany}\\
    \email\small{moritz@pupil-labs.com}\\
  \alignauthor William Patera\\
    \affaddr\small{Pupil Labs UG}\\
    \affaddr\small{Berlin, Germany}\\
    \email\small{will@pupil-labs.com}\\
  \alignauthor Andreas Bulling\\
    \affaddr\small{Max Planck Institute for Informatics}\\
    \affaddr\small{Saarbr\"ucken, Germany}\\
    \email\small{andreas.bulling@acm.org}\\
}

\maketitle

\begin{abstract}
Commercial head-mounted eye trackers provide useful features to customers in industry and research but are expensive and rely on closed source hardware and software. This limits the application areas and use of mobile eye tracking to expert users and inhibits user-driven development, customisation, and extension. In this paper we present Pupil -- an accessible, affordable, and extensible open source platform for mobile eye tracking and gaze-based interaction. Pupil comprises 1) a light-weight headset with high-resolution cameras, 2) an open source software framework for mobile eye tracking, as well as 3) a graphical user interface (GUI) to playback and visualize video and gaze data. Pupil features high-resolution scene and eye cameras for monocular and binocular gaze estimation. The software and GUI are platform-independent and include state-of-the-art algorithms for real-time pupil detection and tracking, calibration, and accurate gaze estimation. Results of a performance evaluation show that Pupil can provide an average gaze estimation accuracy of 0.6 degree of visual angle (0.08 degree precision) with a latency of the processing pipeline of only 0.045 seconds.
\end{abstract}


\section{Introduction}

Eye tracking has been used for over a century to study human behaviour and build insight into cognitive processes~\cite{Wade:Tatler:2005}. Early 20th century eye trackers led to great insight but were large and invasive, and constrained studies to the confines of the laboratory~\cite{Buswell:1935:PeoplePictures,Yarbus:1967:EyeMovements, Wade:2009:Pioneers}. In the second half of 20th century, the first generation of video-based head-mounted eye trackers paved the way for studying visual behaviour during everyday activities outside of the laboratory~\cite{Thomas:1968,Land:1999}.

Recent advances in head-mounted eye tracking and automated eye movement analysis point the way toward unobtrusive eye-based human-computer interfaces that are pervasively usable in everyday life. We call this new paradigm pervasive eye tracking –- continuous eye monitoring and analysis 24/7~\cite{Bulling:10}. The ability to track and analyse eye movements anywhere and anytime will enable new research to develop and understand visual behaviour and eye-based interaction in daily life settings.

Commercially available head-mounted eye tracking systems are robust and provide useful features to customers in industry and research, such as for marketing studies, website analytics, or research studies~\cite{SMI:Glasses,Tobii:Glasses,Ergoneers:Dikablis,ArringtonResearch,ASL:MobileEyeXG}. However, commercial systems are expensive, therefore typically used by specialized user groups, and rely on closed source hardware and software. This limits the potential scale and application areas of eye tracking to expert users and inhibits user-driven development, customisation, and extension.

\begin{figure}[!t]
\centering
\includegraphics[width=0.8\columnwidth]{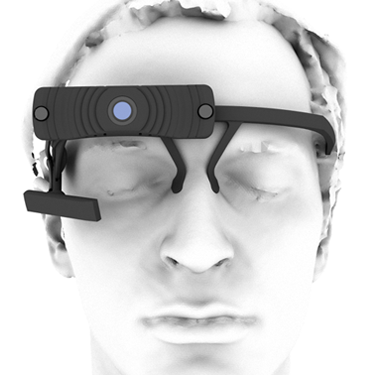}
\caption{Front rendering of the Pupil Pro headset (rev 20) showing the frame, tiltable scene camera and rotatable eye camera.}
\label{fig:pupil_rendering}
\end{figure}

Do-it-yourself (DIY) open source software (OSS) eye trackers have emerged as low cost alternatives to commercial eye tracking systems using consumer digital camera sensors and open source computer vision software libraries~\cite{Babcock:2004,Li:2006:OpenEyes,Ryan:2008:Limbus,Lukander:2013,Mantiuk:2012:DIYEyeTracker,Haytham:2013:Online,Ferhat:14,Topal:2008:HSE}. The DIY/OSS route enables users to rapidly develop and modify hardware and software based on experimental findings~\cite{Bulling:10}.

We argue that affordability does not necessarily align with accessibility. In this paper we define accessible eye tracking platforms to have the following qualities: open source components, modular hardware and software design, comprehensive documentation, user support, affordable price, and flexibility for future changes.

We have developed Pupil, a mobile eye tracking headset and an open source software framework, as an accessible, affordable, and extensible tool for pervasive eye tracking research. In this paper we will explain the design motivation of the system, provide an in depth technical description of both hardware and software, and provide an analysis of accuracy and performance of the system.

\begin{figure*}[!ht]
\centering
\includegraphics[width=2.0\columnwidth]{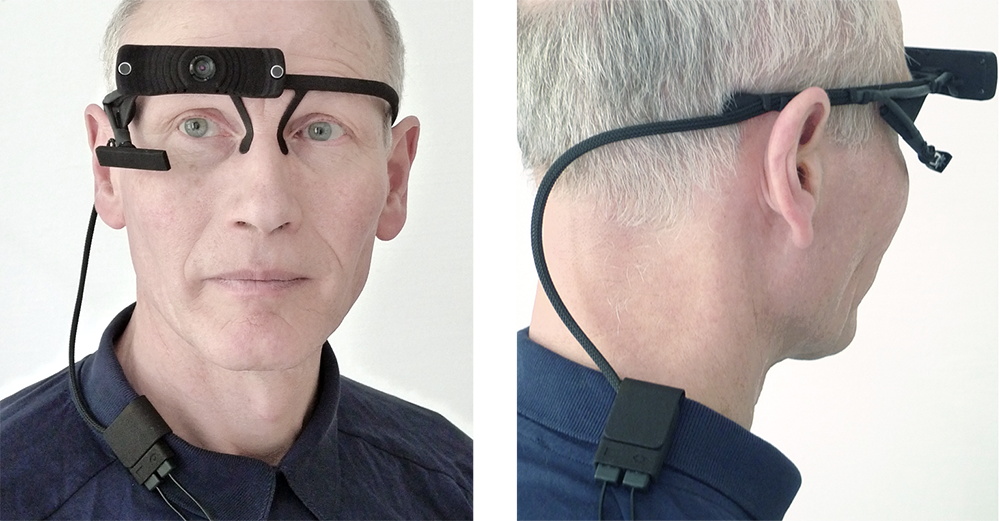}
\caption{Pupil Pro r021. Front view of Pupil Pro (left image), shown with cable clip attached to user's collar. View from behind right (right image) with rotating clip attachment and flexible cable from headset to cable clip. Clip and lightweight and flexible cable enable greater freedom of head movement.}
\label{fig:pupil_on_human}
\end{figure*}

\section{System Overview}
Pupil is a wearable mobile eye tracking headset with one scene camera and one infrared (IR) spectrum eye camera for dark pupil detection.  Both cameras connect to a laptop, desktop, or mobile computer platform via high speed USB 2.0. The camera video streams are read using Pupil Capture software for real-time pupil detection, gaze mapping, recording, and other functions.

\section{System Design Objectives}
In order to design accessible, affordable, and extensible head mounted eye tracking hardware, we made a series of strategic design decisions while satisfying a number of factors to balance ergonomic constraints with performance.

Pupil leverages the rapid development cycle and scaling effects of consumer electronics - USB cameras and consumer computing hardware - instead of using custom cameras and computing solutions.

Pupil headsets are fabricated using Selective Laser Sintering (SLS) instead of established fabrication methods like injection molding. This rapid fabrication process accommodates frequent design changes, comparable to the continuous development of Pupil software.

Modular design principles are employed in both hardware and software enabling modifications by users. Pupil software is open source and strives to build and support a community of eye tracking researchers and developers.

\begin{figure*}[!ht]
\centering
\includegraphics[width=1.8\columnwidth]{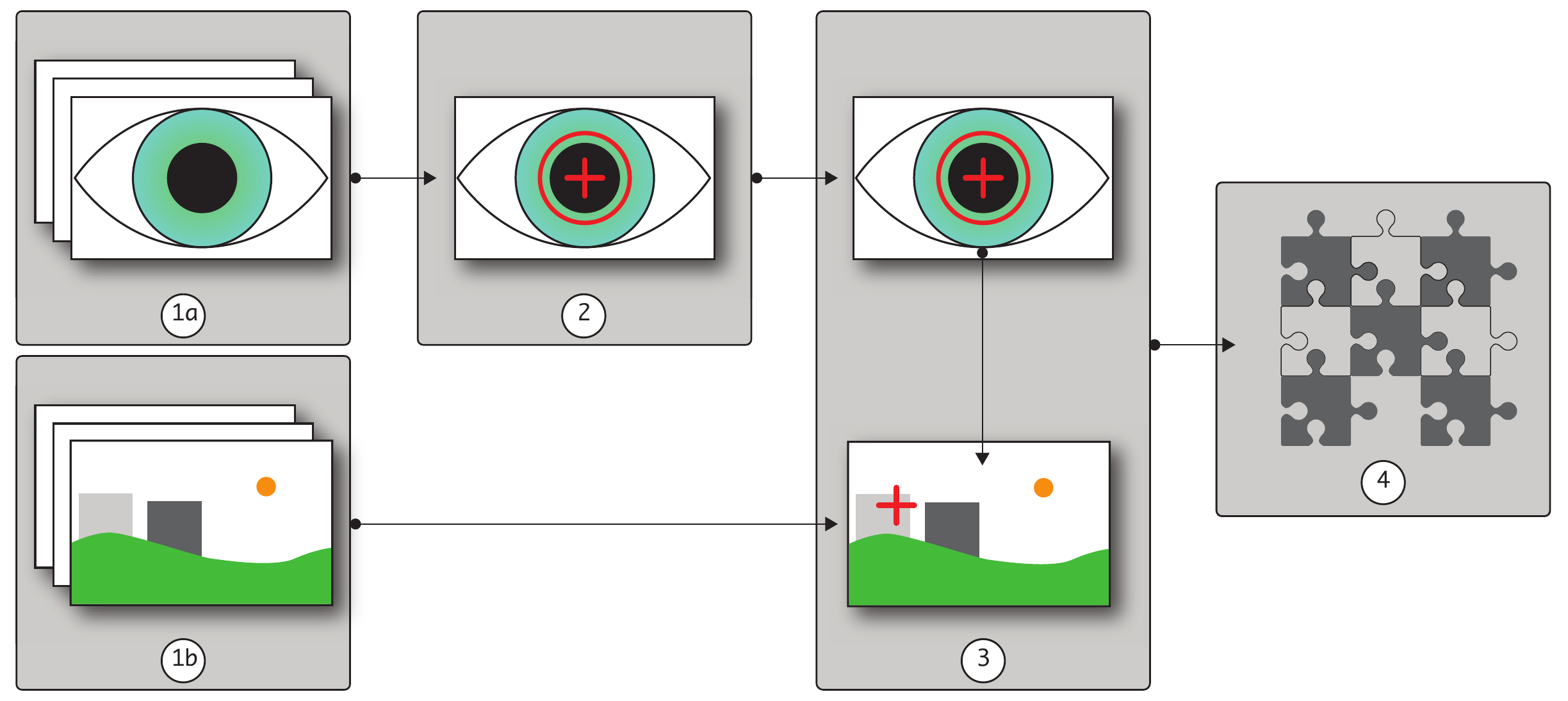}
\caption{Functional overview of Pupil Capture. 1a,1b - Decompress camera image streams for each camera connected. 2 - Detect pupil in eye image. 3 - Map detected pupil position into scene space. 4 - Execute additional functions and plugins (recording, streaming, real-time analysis)}
\label{fig:pupil_capture_flowchart}
\end{figure*}

\vfill

\section{Pupil Headset Design and Hardware}
Factors that critically influence the headset design are: mobility, modularity and customization, minimizing visual obstruction, accommodation of various facial geometries, minimization of headset movement due to slippage and deformation, minimizing weight, durability, and wear comfort.

\subsection{Headset}
The Pupil headset is made up of three modules: frame, scene camera mount, and eye camera mount.

\subsubsection{Frame}
The frame was designed on top of a 3D scan of a human head and was iteratively refined to accommodate physiological variations between users.  We developed a novel design process where we apply specific forces to deform the headset using Finite Element Analysis (FEA) and then print the deformed geometry.  This ensures that cameras align as designed when the headset is worn and results in a well fitting and lightweight (9g) frame.

The frame has two mount points, one for the eye camera mount and one for the scene camera mount.  Cables from each camera are routed through the hollow arm of the frame.  Variations of the frame for binocular eye tracking with four mount points have been made and are currently being tested by the Pupil user community.  Variations of the frame without nose bridge support are available to accommodate users who wear prescription eyeglasses.

\subsubsection{Camera Mounts}
The scene camera mount and eye camera mount interface geometries are open source.  By releasing the mount geometry we automatically document the interface, allowing users to develop their own mounts for cameras of their choice.  All open source mounts are hosted in a Git repository (see links section).

The scene camera mount connects to the frame with a snap fit toothed ratcheting system system that allows for radial adjustment within the users vertical field of vision (FOV) along a transverse axis within a 90 degree range, allowing the scene camera to be adjusted for specific tasks and users.

The eye camera mount is an articulated adjustable arm accommodating variations in users eyes and face geometries. The camera mount attaches to the frame along a snap fit sliding joint. The eye camera orbits on a ball joint that can be fixed by tightening a single screw.

\subsection{Cameras}
Pupil uses USB interface digital cameras that comply with the UVC standard. Other UVC compliant cameras can be used with the system as desired by the user. The Pupil headset can be used with other software that supports the UVC interface. Pupil can be easily extended to use two eye cameras for binocular setups and more scene cameras as desired.

\subsubsection{Eye Camera}
We use a small and lightweight eye camera to reduce the amount of visual obstruction for the user and keep the headset lightweight. The current eye camera package size for Pupil Pro is 10x45x7 mm. The eye camera can capture at a maximum resolution of 800x600 pixels at 30Hz. Using an IR mirror (``hot mirror'') was considered as a strategy to further reduce visual obstruction and conceal cameras but was ultimately dismissed as it would introduce more degrees of freedom in the setup that could negatively affect performance, ergonomics, and modularity. Furthermore, hot mirror setups are susceptible to failure in environments with high amounts of IR light (like sunlight).

Pupil uses the ``dark pupil'' detection method (see Pupil Detection Algorithm Overview). This requires the eye camera to capture video within a specific range of the IR spectrum. The eye camera uses an IR bandpass filter and a surface mounted IR LED at 860nm wavelength to illuminate the user's eye.

\subsubsection{Scene Camera}
The scene camera is mounted above the user's eye aligning the scene camera optics with the user's eye along a sagittal plane. The scene camera faces outwards to capture a video stream of a portion of the user’s FOV at 30Hz. The scene camera lens has a 90 degree diagonal FOV. The scene camera is not only high resolution (max resolution 1920x1080 pixels), but also uses a high quality image sensor. This is very advantageous for further computer vision and related tasks performed in software.

\subsection{Computing Device}
The Pupil eye tracking system works in conjunction with standard multipurpose computers: laptop, desktop, or tablet. Designing for user supplied recording and processing hardware introduces a source for compatibility issues and requires more setup effort for both users and developers. However, enabling the user to pair the headset with their own computing platform makes Pupil a multipurpose eye tracking and analysis tool. Pupil is deployable for lightweight mobile use as well as more specialized applications like: streaming over networks, geotagging, multi-user synchronization; and computationally intensive applications like real time 3D reconstruction and localization.

\begin{figure*}[!ht]
\includegraphics[width=2.0\columnwidth]{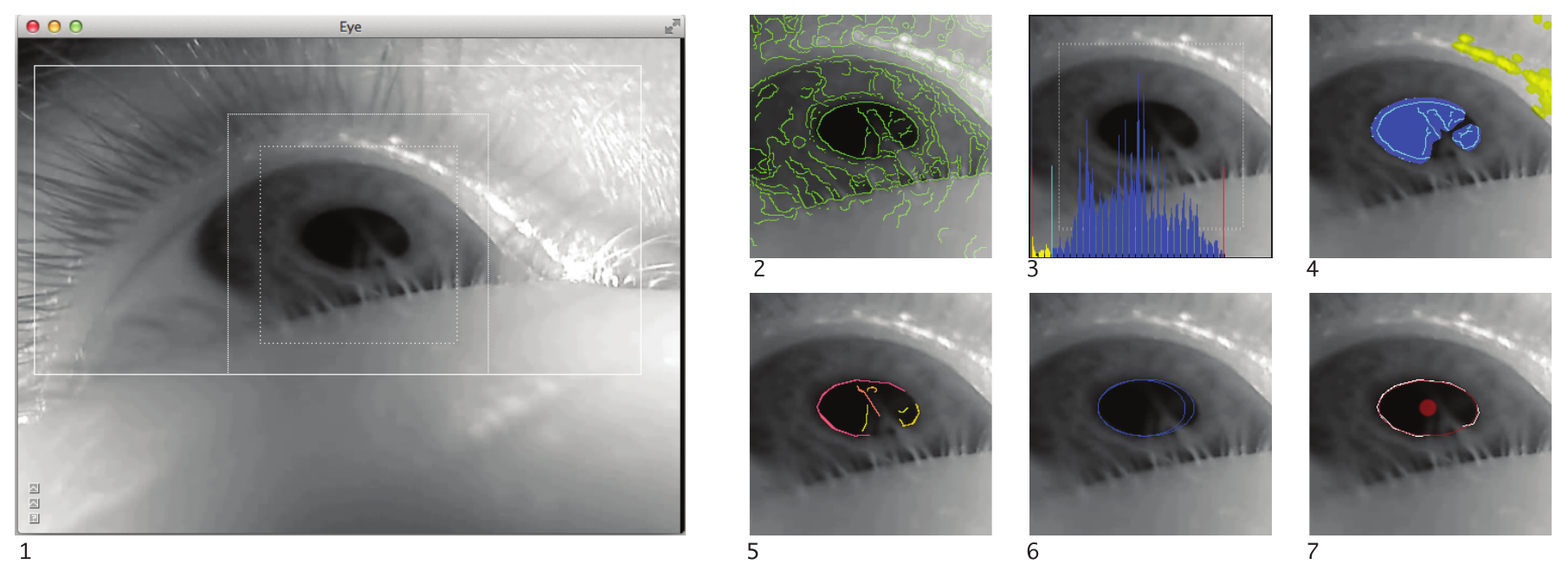}
\caption{Vizualization of pupil detection algorithm. 1) Eye image converted to grayscale, user region of interest (white stroke rectangle), and initial estimation of pupil region (white square and dashed line square.) 2) Canny edge detection (green lines.) 3) Define ``dark'' region as offset from lowest spike in histogram within eye image. 4) Filter edges to exclude spectral reflections (yellow.) and not inside ``dark'' areas (blue) 5) Remaining edges extracted into contours using connected components and split into sub-contours based on curvature continuity criteria (multi colored lines). 6) Candidate pupil ellipses (blue) are formed using ellipse fitting 7) Final ellipse fit found through an augmented combinatorial search ( finally ellipse with center in red) - supporting edge pixels drawn in white.}
\label{fig:pupil_detection_algorithm}
\end{figure*}

\section{Pupil Software}
Pupil software is open source code (CC-BY-NC-SA License) written to be readable, extendable, robust and efficient.  Pupil software is divided into two main parts, Pupil Capture and Pupil Player.  Pupil Capture runs in real-time to capture and process images from the two (or more) camera video streams.  Pupil Player is used to playback and visualize video and gaze data recorded with Pupil Capture. Source code is written mostly in Python~\cite{Python:2.7} and modules are written in C where speed is a concern. Pupil software and can be run from source on Linux, MacOS (10.8 or higher), and Windows or executed as a bundled double click application on Linux and MacOS.

Pupil software depends on open source libraries: OpenCV, FFMPEG, NumPy, PyOpenGL, AntTweakBar, ZeroMQ, and GLFW~\cite{OpenCV:Library,NumPy:Library,Hintjens:2013:ZeroMQ,FFMPEG:2013,GLFW:2013,ATB:2013}.

\begin{figure*}
\centering
\includegraphics[width=2.0\columnwidth]{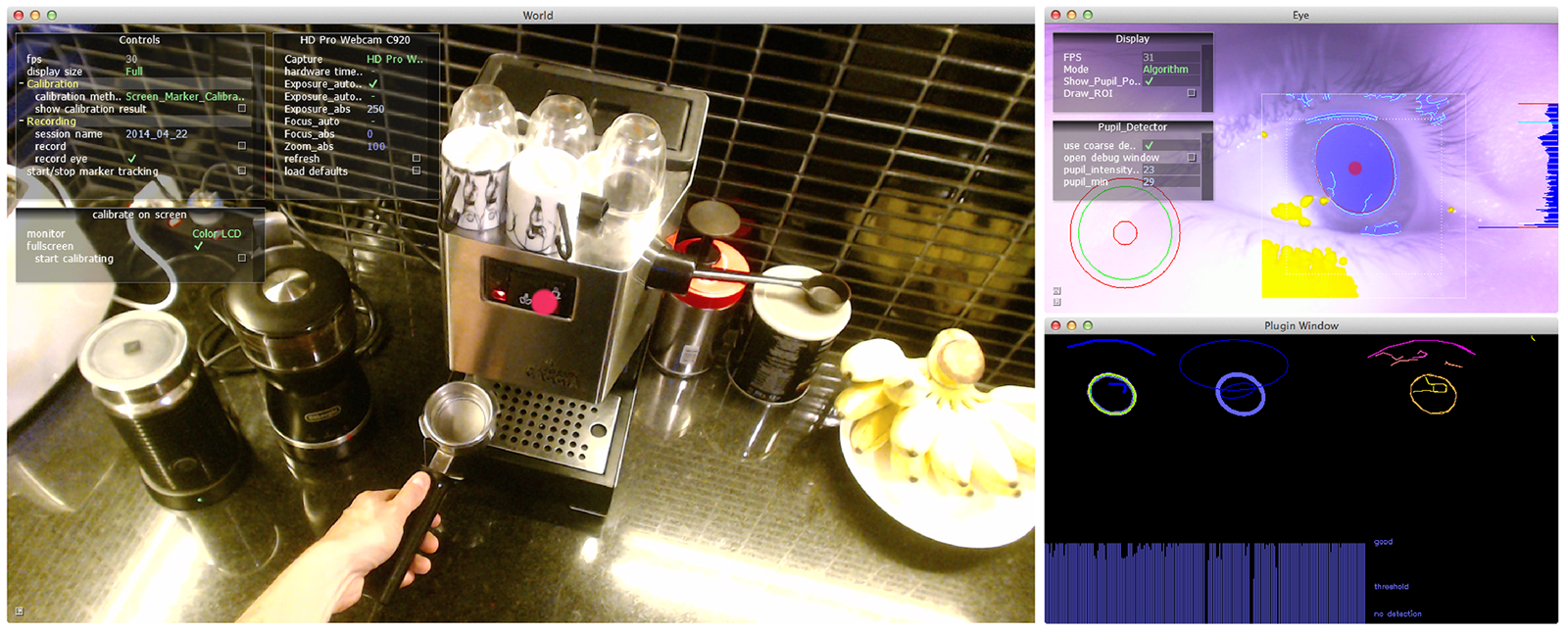}
\caption{Screen capture of Pupil Capture. World) Window displays real-time video feed of the user's FOV from the scene camera along with GUI controls for the world camera and plugins. Red circle is the gaze position of the user. Eye) Window displays real-time video feed of the user's eye from the eye camera with GUI controls for the eye camera and plugins for the eye process. Plugin Window) plugins can also spawn their own windows. Shown here is a plugin to visualize the pupil detection algorithm.}
\label{fig:pupil_capture_screenshots}
\end{figure*}

\subsection{Key Functional Components Overview}
This section provides an overview of key functions of Pupil Capture Software.

\subsubsection{Pupil Detection Algorithm}
\label{subsec:pupil-detection-algorithm}
The pupil detection algorithm locates the dark pupil in the IR illuminated eye camera image. The algorithm does not depend on the corneal reflection, and works with users who wear contact lenses and eyeglasses. The pupil detection algorithm is under constant improvement based on feedback collected through user submitted eye camera videos. Here we provide a description of our default pupil detection algorithm.

The eye camera image is converted to grayscale. The initial region estimation of the pupil is found via the strongest response for a center-surround feature as proposed by  Swirski et al.~\cite{Swirski:12} within the image (See Figure:~\figref{fig:pupil_detection_algorithm}.)

Detect edges using Canny~\cite{Canny:86} to find contours in eye image. Filter edges based on neighboring pixel intensity. Look for darker areas (blue region). Dark is specified using a user set offset of the lowest spike in the histogram of pixel intensities in the eye image.
Filter remaining edges to exclude those stemming from spectral reflections (yellow region).
Remaining edges are extracted into into contours using connected components ~\cite{Suzuki:85}.
Contours are filtered and split into sub-contours based on criteria of curvature continuity.
Candidate pupil ellipses are formed using ellipse fitting~\cite{Fitzgibbon:95} onto a subset of the contours looking for good fits in a least square sense, major radii within a user defined range, and a few additional criteria.
An augmented combinatorial search looks for contours that can be added as support to the candidate ellipses.
The results are evaluated based on the ellipse fit of the supporting edges and the ratio of supporting edge length and ellipse circumference (using Ramanujan’s second approximation~\cite{Ramanujan:62}). We call this ratio ``confidence''.
If the best result’s confidence is above a threshold the algorithm reports this candidate ellipse as the ellipse defining the contour of the pupil. Otherwise the algorithm reports that no pupil was found.

\begin{figure}
\includegraphics[width=1.0\columnwidth]{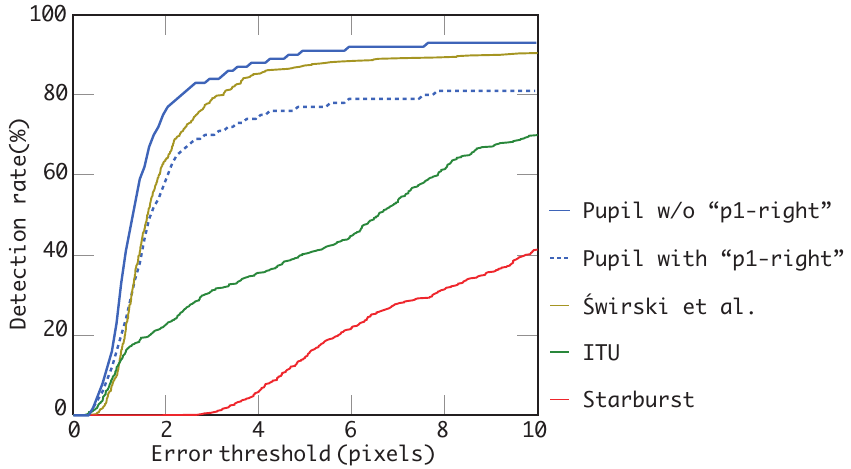}
\caption{Comparison of pupil detection rate for Pupil's algorithm, the stock algorithm proposed by Swirski et al., the ITU gaze tracker and Starburst. (Figure data for algorithms other than Pupil used with permission of Swirski.)}
\label{fig:algorithm_evaluation}
\end{figure}

Figure~\figref{fig:algorithm_evaluation} shows a performance comparison between Pupil's pupil detection algorithm, the stock algorithm proposed by Swirski et al., the ITU gaze tracker and Starburst on the benchmark dataset by Swirski et al.~\cite{Swirski:12}. As error measure we used the Hausdorff distance between the detected and hand-labeled pupil ellipses~\cite{Swirski:12}. We additionally conducted a test excluding the dataset p1-right, that contains eye images recorded at the most extreme angles, as those do not occur using Pupil hardware. As can be seen from the Figure, Pupil without p1-right compares favourably to all other approaches. With an error threshold of 2 pixels Pupil achieves a detection rate of 80\%; at 5 pixels error detection rate increases to 90\%.

\subsubsection{Gaze Mapping}
Mapping pupil positions from eye to scene space is implemented with a transfer function consisting of two bivariate polynomials of adjustable degree. The user specific polynomial parameters are obtained by running one of the calibration routines:
\begin{itemize}
\item Screen Marker Calibration - 9 Point animated calibration method.
\item Manual Marker Calibration - Uses a concentric circle marker that can be moved freely within a user's FOV. One marker pattern is used to collect samples for calibration, another marker pattern is used to stop the calibration process.
\item Natural Features Calibration - Uses natural features within the scene. Features are defined by clicking on a salient point within the world window. Features are tracked using optical flow.
\item Camera Intrinsic Calibration - Used to calculate camera intrinsics with an 11x7 asymmetric circle grid pattern.
\end{itemize}

Calibration and mapping functions are abstracted and the underlying models can easily be modified and replaced if needed.

\subsubsection{Surface Detection}
Pupil Capture can detect planar reference surfaces in the scene using a set of 64 markers. Gaze positions are then mapped into the reference surface coordinate system using homographic transformations between the scene camera plane and reference surface plane. See a video demonstration of planar reference in links.

\subsubsection{Recording}
Pupil Capture can record the scene and eye videos, associated frame timestamps, detected pupil and gaze position data, and additional user activated plugin data.

\newlength{\hcolw}
\setlength{\hcolw}{0.12\columnwidth}

\begin{table}[!ht]
    \begin{scriptsize}
    \centering
    \begin{tabular}{|p\hcolw|p\hcolw|p\hcolw|p\hcolw|p\hcolw|p\hcolw|}
    \hline
    {\bf gaze x} & {\bf gaze y} & {\bf pupil x} & {\bf pupil y} & {\bf timestamp} & {\bf confidence} \\
    \hline
    0.585903 & 0.344576 & 0.538961 & 0.473854 & 0.139290 & 0.97686 \\
    \hline
    \end{tabular}
    \end{scriptsize}
  \caption{Example row of data saved from Pupil Capture. Data is saved as 64 bit floating point shown here with 6 significant digits.}
  \label{tab:pupil_sample_data_row}
\end{table}

\subsubsection{Streaming}
Pupil Capture can send real-time gaze and pupil information as well as plugin generated data via ZMQ to other applications and network enabled devices.

\begin{figure*}[!ht]
\centering
\includegraphics[width=2.0\columnwidth]{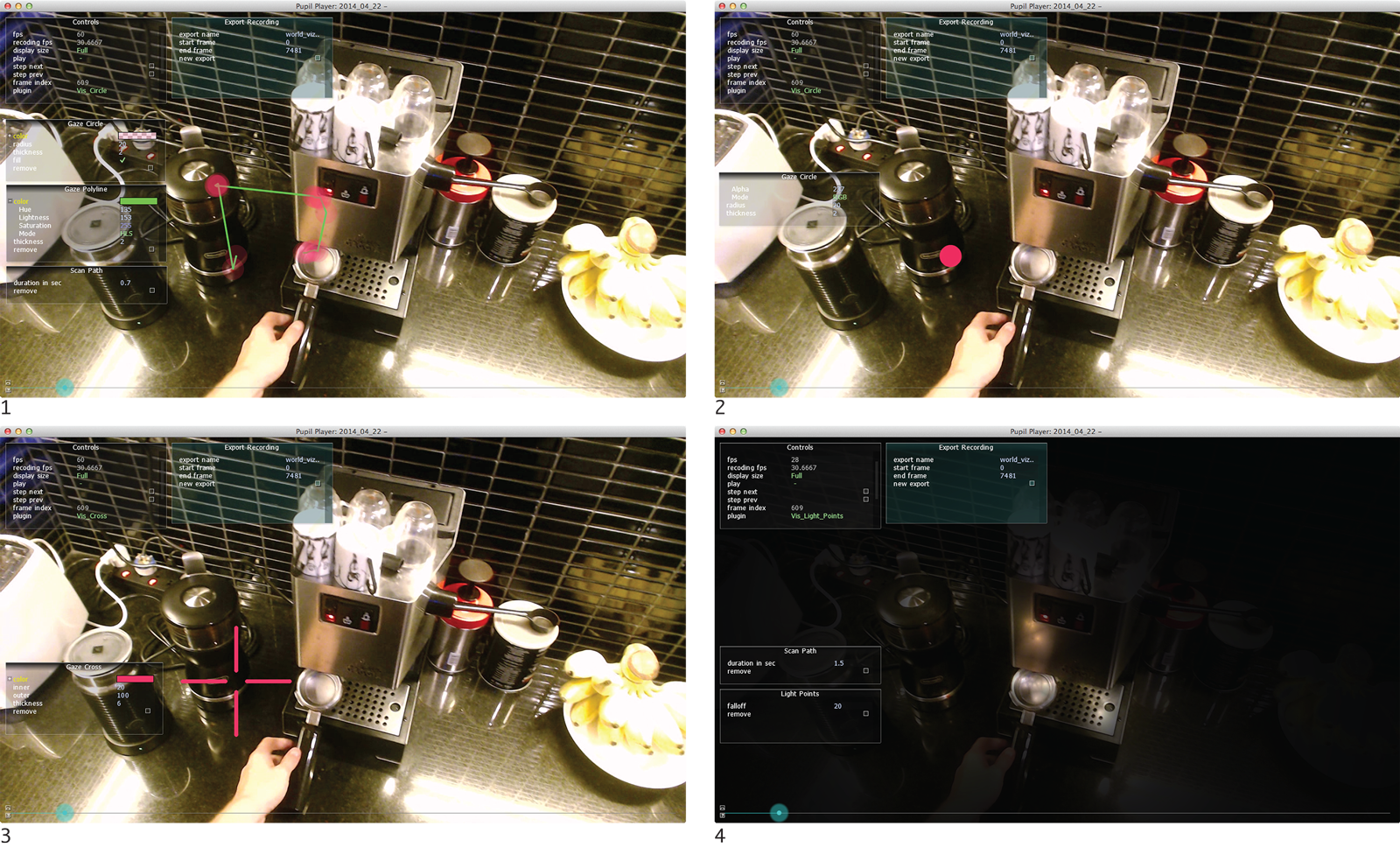}
\caption{Screen captures of Pupil Player demonstrating four different visualization methods of the same scene at the same point in time. 1) Screen capture of Pupil Player showing a visualization of gaze data and scene video recorded with Pupil Capture. Three plugins used additively to generate the visualization: i) Scan Path - shows gaze positions within duration set by the user and uses optical flow to compensate for ego-motion. ii) Gaze Circle - translucent red circles drawn for each gaze position within the range set by the Scan Path plugin. iii) Gaze Polyline - green line that connects the gaze positions in temporal order. 2) Gaze Circle visualization current data point as red filled circle. 3) Cross visualization - current data point as red cross mark. 2c) Gaze Polyline - gaze points within temporal duration of Scan Path are shown as green polyline. 4) Light Point Transform - gaze points within temporal duration of Scan Path are shown as white white points }
\label{fig:pupil_player_vis_examples}
\end{figure*}

\subsection{Pupil Capture Implementation}
Each process has the ability to launch plugins. Even a standard feature, like recording a video, is abstracted as a plugin. This level of abstraction and modularity allows for users to develop their own tools even for low level functionality.  Plugins are loaded automatically and are accessible through the graphical user interface (GUI).

\subsubsection{World Window}
The world window can be considered the main control window in the Pupil Capture environment.  The world window displays the video stream from the scene camera, camera controls for the scene camera, and launches plugins like calibration or recording (see plugins section for more detail).

\subsubsection{Eye Window}
The eye window displays the video stream from the eye camera, camera controls for the eye camera, and launches plugins for pupil detection and visualization of pupil detection algorithms.

\subsubsection{Plugin Structure}
In Pupil Capture plugins can be launched in either world or eye processes. The number of plugins that can be created is limitless and much of the functionality of the eye and world processes has been abstracted as plugins. The modular structure makes it easy to for users to test out different methods at runtime and for developers to extend software without breaking existing functionality. Plugins have the ability to create their own GUI within the process window, access to shared memory like gaze data, and even the capacity spawn their own windows.

\section{Performance Evaluation}

\subsection{Spatial Accuracy and Precision}
As performance metrics for spatial accuracy and precision of the system we employ the metrics defined in COGAIN Eye tracker accuracy terms and definitions~\cite{COGAIN}

``{\bf Accuracy} is calculated as the average angular offset (distance) (in degrees of visual angle) between fixations locations and the corresponding locations of the fixation targets.''

``{\bf Precision} is calculated as the Root Mean Square (RMS) of the angular distance (in degrees of visual angle) between successive samples to \verb|xi, yi to xi+1,yi+1|.''

It is important to note that the test is done in an indoor environment with no direct sunlight exposure. Pupil detection performance in an IR intense environment is more challenging and needs to be evaluated separately. The authors believe that perfect detection in IR intense environments remains an unsolved problem of IR-based pupil-detection.

Our accuracy and precision test data was obtained by the following procedure:

A subject wearing the Pupil Pro eye tracker sits approximately 0.5m away from a 27 inch computer monitor. The eye tracker is calibrated using the standard 9 point screen marker based calibration routine running in full screen mode. After calibration the subject is asked to fixate on a marker on the monitor. The marker samples 10 random sites for 1.5 seconds and then revisits each of the 9 calibration sites. While the marker is displayed gaze samples and the position of the marker detected in the scene camera via software are collected. The data is then correlated into gaze point and marker point pairs based on minimal temporal distance.

Gaze point marker point pairs with a spatial distance of more than 5 degrees angular distance are discarded as outliers as they do not correlate to system error but human error (no fixation, blink)~\cite{TobiiTest:Whitepaper}.

Accuracy and precision are then calculated in scene camera pixel space and converted into degrees of visual angle based on camera intrinsics. For this particular test we used a scene camera with 90 degrees (diagonal) field of view and 1280x720 pixel resolution. This results in a ratio of 16.32 pixels per degree of visual angle.

This procedure was repeated with eight different subjects to reflect a more diverse pool of physiologies. The test was conducted with a Pupil Pro eye tracker revision 021. The test used Pupil Capture software, version 0.3.8 running on Mac OS. The test routine is part of Pupil Capture releases, starting with version 0.3.8 and available to all users.

\subsubsection{Error Sources}
The accuracy and precision test expose the gaze estimation error. It is composed of a set of individual error sources. Individual attribution is not in the scope of this test. (authors estimates in italic)

\begin{itemize}
\item Oculomotor noise (tremor, microsaccades, drift)~\cite{COGAIN}
\item Subject human error during test (not looking at the marker, blinks) \emph{most filtered out by 5 deg rule}
\item Limitations of human fixation accuracy (0.5deg)~\cite{Solso:1996} \emph{in light of the achived accuracy this should be regarded as a major factor}
\item Movement and deformation of the headset resulting in changes of camera positions relative to the eye post calibration. \emph{can be disregarded in this test}
\item Image noise in the sensor, interpolation and compression artifacts. \emph{disregarded if assumed to be Gaussian this would show in precision being worse}
\item Pupil detection error by the algorithm. \emph{See algorithm evaluation}
\item Gaze Mapping error based on shortcomings of the mapping model \emph{we believe this to be a factor}
\item Gaze Mapping error based on suboptimal fitting parameters from flawed calibration procedure design. \emph{a factor}
\item Gaze Mapping error based on suboptimal fitting parameters obtained from  flawed calibration procedure execution. \emph{small}
\item Parallax error due to change of distance between subject of gaze target. \emph{Can be a big factor in real world applications but can be disregarded in the test scenario}
\item Error in test reference data from suboptimal center detection of the marker. \emph{can be disregarded - detection is based on many redundant pixels and very robust and accurate}
\item Error in the test reference data from temporal discrepancy of the sample pair. \emph{Since we have little movement in head and marker during collection of sample we can disregard this.}
\item Conversion error based on flawed camera intrinsics data. \emph{can be disregarded}
\end{itemize}

\subsubsection{Results}
\begin{figure}[!t]
\includegraphics[width=1.0\columnwidth]{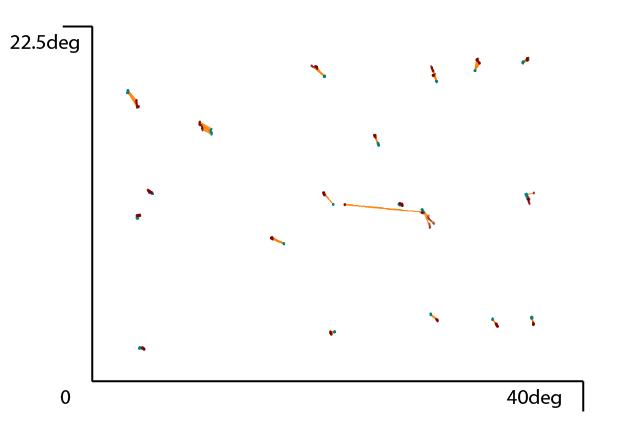}
\caption{Results of accuracy test for one user. Marker target in green. Gaze point in red. Correspondace error in orange. Notice sample point with big error due to subject fixation failure at the begining of the test.}
\label{fig:accuracy_test}
\end{figure}
Preliminary results:

\begin{itemize}
  \item Under ideal conditions we get 0.6 deg of accuracy.
  \item Precision is at 0.08 deg.
\end{itemize}

\begin{figure*}[!ht]
\centering
\includegraphics[width=2.0\columnwidth]{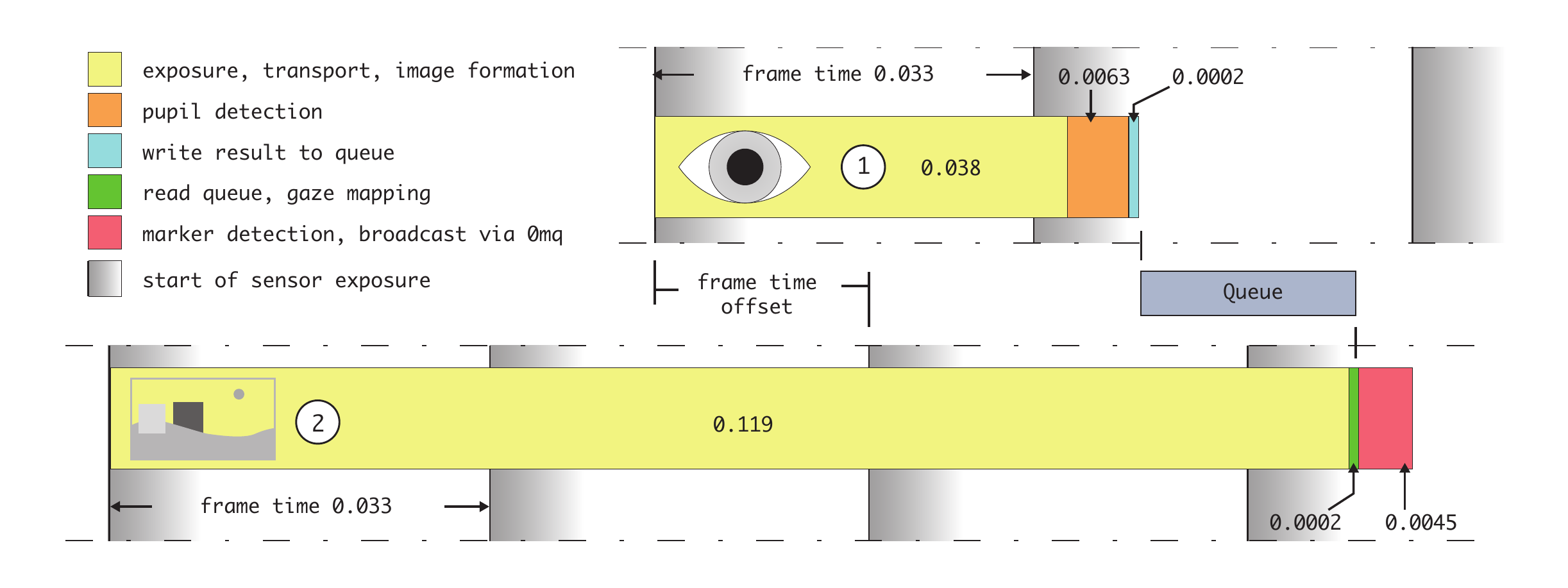}
\caption{System latency diagram shows 1) eye process and 2) scene processes and times between sensor exposure and key points in the processing pipeline. Gray bars represent video streams of each process with vertical divisions as individual image frames within the streams. Color bars for key functions annotated with processing times.}
\label{fig:system-latency}
\end{figure*}

\subsection{Temporal Accuracy, Latency and Precision}
A second critical aspect are the temporal characteristics of the pupil eye tracking system.

\subsubsection{Stream Synchronization}
Timestamping is crucial for stream stream synchronization because data is obtained from two independent free running video sources. Additionally timestamps are used for correlation with additional external experiment or sensor data. Thus we strive to obtain a timestamp that is closest to the timepoint of data collection (in our case camera sensor exposure).

Pupil capture has two image timestamp implementations:

When the imaging camera is known to produce valid hardware timestamps, Pupil Capture uses these hardware image timestamps. These timestamps have high precision and accuracy as they are taken at the beginning of sensor exposure by the camera and transmitted along with the image data. The hardware timestamp accuracy exceeds our measurement capabilities.

The variation of exposure times (jitter) reported by the hardware timestamps we measure by calculating the standard deviation of 1400 successively take frame times. It is 0.0004s for the world camera and 0.0001s for the eye camera.

Currently hardware timestamping is implemented in the Linux version of Pupil Capture and supported by both cameras of the Pupil Pro headset revision 020 and up.

When Pupil Capture runs on an OS without timestamp video driver support or does not recognize the attached cameras as verified hardware timestamp sources, Pupil Capture uses software timestamps as a fallback. The image frame timestamp is then taken when the driver makes the image frame available to the software.

Software timestamps are by nature of worse accuracy and precision, as they is taken after exposure, readout, image transfer, and decompression of the image on the host side. These steps take an indeterministic amount of time, which makes it impossible to accurately estimate time of exposure. Accuracy and precision depend on the camera and video capture driver.

Taking hardware timestamps as ground truth we measured a software timestamp offset of +0.119s (world) and +0.038s (eye) with a standard deviation of 0.003 (world) and 0.002s (eye) on a Linux OS test machine. These results clearly underline that hardware timestamps should always be used when temporal accuracy and precision are required.

\subsubsection{System Latency}
For real-time applications the full latency of Pupil hardware and software is of great concern. We obtain processing times of functional groups in the Pupil hardware and software pipeline by calculating the time between sensor exposure and key points in the workflow. These temporal factors in signal delay are presented in Figure~\figref{fig:system-latency}. All measurement were taken using Pupil Pro headset rev022 connected to a Lenovo X201 laptop running Ubuntu 12.04.

It should be noted that in real-time applications synchronicity of data is sacrificed for recency of data. The eye process pipeline is about 1/3 the latency of the world process pipeline. Therefore, we choose to broadcast the eye information as soon as it becomes available instead of waiting for the temporally closest scene image to become available.

Using the most recent data only makes sense for real-time applications. No sacrifices are made for any after-the-fact correlation of data employed by calibration, testing, or playback of recorded data. Furthermore, this approach does not prevent the user from obtaining accurate temporal correlation in real-time or after-the-fact applications.

With this approach we can characterize the system latency for the eye pipeline, world pipeline separately:
\begin{itemize}
 \item Total latency of the eye processing pipeline from start of sensor exposure to availability of pupil position: 0.045s (measured across 1400 samples with a standard deviation of 0.003sec)
 \item  Total latency of the world pipeline including the eye measurement from start of sensor exposure to broadcast of pupil, gaze, and reference surface data via network: 0.124 sec (measured across 1200 samples with a standard deviation of 0.005 sec)
\end{itemize}

\subsection{Minimum Hardware Requirements}
The Pupil eye tracking system works with a traditional multipurpose computer - laptop, desktop, or tablet. It is therefore important to determine the minimum hardware specifications required to run Pupil Capture software in real-time. We tested the performance using a 11 inch Macbook Air (2010 model) with 2gb or RAM and an Intel Core2Duo SU9400 dual core CPU. The software version used was Pupil Capture v0.3.8. The OS used on the machine specified above was Ubuntu 13.10.

Our performance test demonstrates that the system's dual CPU load never went above 90 percent, using the above hardware running Pupil Capture in recording mode, pupil detection at 30 fps and the world camera capture at 24 fps.

The hardware setup was selected because it represents a portable computing platform with limited computing power compared to most contemporary consumer multipurpose computers.  Any computer with a Intel ``i'' series processor or equivalent will have sufficient CPU resources, when comparing CPU benchmarks.

Pupil Capture relies on several libraries to do video decompression/compression, image analysis, and display with platform specific implementations and efficiencies. Therefore, our test using a Linux distribution can not be generalized to Windows or MacOS. We found that requirements were similar for MacOS 10.8 and above. We did not establish Windows hardware requirements at the time of writing.

\section{Discussion}

In order to further advance in eye tracking and to support the pervasive eye tracking paradigm, we will require accessible, affordable, and extensible eye tracking tools. We have developed Pupil as a contribution to the eye tracking research community. Pupil is already used in a wide range of disciplines and has developed a community of researchers and developers. Current limitations to the system are parallax error and tracking robustness in IR rich environments. Both Pupil software and hardware are under active development. Future developments will focus on hardware and software in parallel. The next big steps planned for Pupil are to improve mobility, implement real-time pose tracking and scene mapping, simplify user experience, and improve pupil tracking. 

\section{Links}
\begin{enumerate}
  \item Pupil mobile eye tracking headset: \url{http://pupil-labs.com/pupil}
  \item Pupil open source code repository: \url{http://github.com/pupil-labs/pupil}
  \item Pupil Capture and Pupil Player software application bundles: \url{https://github.com/pupil-labs/pupil/releases}
  \item Pupil User Guide: \url{https://github.com/pupil-labs/pupil/wiki/User-Guide}
  \item Pupil Developer Guide: \url{https://github.com/pupil-labs/pupil/wiki/Developer-Guide}
  \item Pupil user group forum: \url{http://groups.google.com/forum/#!forum/pupil-discuss}
  \item Pupil Labs blog: \url{http://pupil-labs.com/blog} 
  \item Pupil open source headset mount repository \url{https://code.google.com/p/pupil/source/browse/?repo=hardware}
  \item Video demonstration - Planar reference surface tracking: \url{http://youtu.be/bmqDGE6a9kc}
\end{enumerate}

\balance

\bibliographystyle{acm-sigchi}
\bibliography{references}
\end{document}